\documentclass[journal]{IEEEtran} %
\usepackage{amsmath,amsfonts} 
\usepackage{algorithm}
\usepackage{algorithmic}
\usepackage{array}
\usepackage[caption=false,font=normalsize,labelfont=sf,textfont=sf]{subfig}
\usepackage{textcomp}
\usepackage{stfloats} %
\usepackage{url} 
\usepackage{verbatim}
\usepackage{graphicx} %
\usepackage{cite} %
\usepackage{multirow}
\usepackage{makecell}
\usepackage{threeparttable}
\usepackage{float} %
\usepackage[normalem]{ulem} 
\useunder{\uline}{\ul}{}

\begin{document}
\title{GraSS: Contrastive Learning with Gradient Guided Sampling Strategy for Remote Sensing Image Semantic Segmentation}

\author{Zhaoyang Zhang,~
        Zhen Ren,~
        Chao Tao,~
        Yunsheng Zhang,
        Chengli Peng,
        and~Haifeng Li*~
\thanks{The work is financial supported by the Major Program/Open Project of Xiangjiang Laboratory (No. 22XJ01010), The National Natural Science Foundation of China (No. 42171376, 41771458), The Natural Science Foundation of Hunan for Distinguished Young Scholars (No. 2022JJ10072); and the High-Performance Computing Center of Central South University. (Corresponding author: H.F. Li, lihaifeng@csu.edu.cn)}
\thanks{Z. Zhang, Z. Ren, C. Tao, Y. Zhang, C. Peng, and H. Li are with the School of Geosciences and Info-Physics, Central South University, Changsha 410083, China. Z. Zhang and H. Li are also with the Xiangjiang Laboratory, Changsha 410205, China.}
}

\markboth{Journal of \LaTeX\ Class Files,~Vol.~13, No.~9, June~2023} 
{GraSS for Remote Sensing Image Semantic Segmentation} 

\maketitle

\begin{abstract}
Self-supervised contrastive learning (SSCL) has achieved significant milestones in remote sensing image (RSI) understanding. Its essence lies in designing an unsupervised instance discrimination pretext task to extract image features from a large number of unlabeled images that are beneficial for downstream tasks. However, existing instance discrimination based SSCL suffers from two limitations when applied to the RSI semantic segmentation task: 1) Positive sample confounding issue, SSCL treats different augmentations of the same RSI as positive samples, but the richness, complexity, and imbalance of RSI ground objects lead to the model actually pulling a variety of different ground objects closer while pulling positive samples closer, which confuse the feature of different ground objects. 2) Feature adaptation bias, SSCL treats RSI patches containing various ground objects as individual instances for discrimination and obtains instance-level features, which are not fully adapted to pixel-level or object-level semantic segmentation tasks. To address the above limitations, we consider constructing samples containing single ground objects to alleviate positive sample confounding issue, and make the model obtain object-level features from the contrastive between single ground objects. Meanwhile, we observed that the discrimination information can be mapped to specific regions in RSI through the gradient of unsupervised contrastive loss, these specific regions tend to contain single ground objects. Based on this, we propose contrastive learning with Gradient guided Sampling Strategy (GraSS) for RSI semantic segmentation. GraSS consists of two stages: 1) the instance discrimination warm-up stage to provide initial discrimination information to the contrastive loss gradients, 2) the gradient guided sampling contrastive training stage to adaptively construct samples containing more singular ground objects using the discrimination information. Experimental results on three open datasets demonstrate that GraSS effectively enhances the performance of SSCL in high-resolution RSI semantic segmentation. Compared to eight baseline methods from six different types of SSCL, GraSS achieves an average improvement of 1.57\% and a maximum improvement of 3.58\% in terms of mean intersection over the union. Additionally, we discovered that the unsupervised contrastive loss gradients contain rich feature information, which inspires us to utilize gradient information more extensively during model training to attain additional model capacity. The source code is available at https://github.com/GeoX-Lab/GraSS.
\end{abstract}


\IEEEpeerreviewmaketitle

\vspace{-5mm}
\section{Introduction}

\begin{figure}
\centering
\includegraphics[width=8.5cm]{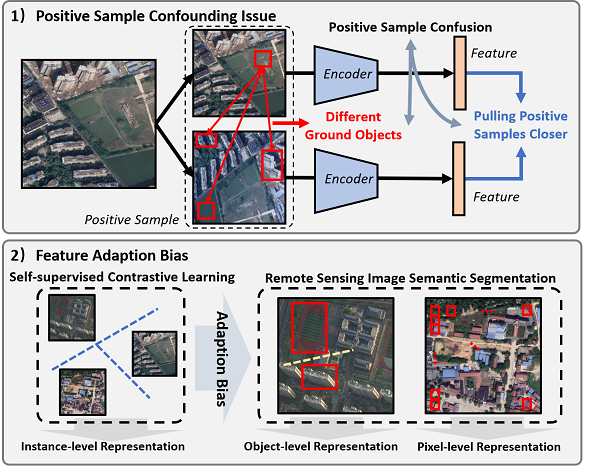}
\vspace{-3mm}
\caption{Example of the positive sample confounding issue and feature adaption bias in self-supervised contrastive learning for RSI semantic segmentation. 1) The richness, and complexity of RSI ground objects can result in positive samples containing different ground objects, self-supervised contrastive learning (SSCL) lead to positive sample confounding issue when pulling these positive samples closer in the feature space. 2) SSCL treats RSI patches containing various ground objects as individual instances for discrimination, resulting in feature representations at the instance level, it introduces a feature adaptation bias when applied to semantic segmentation tasks that require pixel-level or object-level features.}
\label{fig_p12}
\end{figure}

\IEEEPARstart{S}{elf}-supervised contrastive learning has achieved significant success in various downstream tasks such as remote sensing image (RSI) scene classification\cite{RS_SSL2,RS_SSL1,STICL,LasT,better}, hyperspectral image classification\cite{zhao2022hyperspectral,zhang2022hyperspectral,zhang2023single,zhang2021topological,zhang2023language,zhang2022graph}, object detection\cite{DetCo,InsLoc}, change detection\cite{SSL_CD,SSL_CD2}, and semantic segmentation\cite{GLCNet,IndexNet,FALSE}. Its core idea is to learn effective image representations by designing an unsupervised instance discrimination pretext task \cite{TOV,SimCLR,SimCLRv2,RS-MetaNet,MoCo,MoCov2,MoCov3,misra2020self,RS_SSL_Re}.

However, there are two limitations of the unsupervised instance discrimination pretext task when applied to the task of RSI semantic segmentation that requires capturing features of ground objects\cite{GLCNet,IndexNet,FALSE,DenseCL,unsupervised_dense}. First, the positive sample confounding issue. Positive sample confounding issue is one aspect of the sample confounding issue (SCI)\cite{FALSE,Ad_SSL}. Due to the richness, complexity, and imbalance of ground objects contained in RSIs\cite{zhu2020map,SCAttNet,FALSE}, the sample confounding issue in self-supervised contrastive learning models manifests in two aspects: The first aspect is that for negative samples, the self-supervised contrastive learning model treats augmentations of all different images as negative samples, which leads the model to inevitably push away negative samples that contain the same ground objects as the anchor sample, we call this the negative sample confounding issue, which is also often referred to as the false negative sample issue\cite{FN_1,FN_2,FALSE}. The second aspect is that for positive samples, since RSIs contain a variety of ground objects, the model actually pulls a variety of different ground objects in positive samples closer while pulling positive samples closer, which makes the model confuse the features of different ground objects, we call this the positive sample confounding issue (as shown in Fig. \ref{fig_p12}). The positive sample confounding issue undermines the identity assumption of self-supervised contrastive learning\cite{SSL_Re,ID_SSL_Ad,Pre_SSL,UnderstandCL}, which is the focus of this study. Second, the feature adaptation bias. Self-supervised contrastive learning treats RSI patches containing various ground objects as individual instances, focusing more on the relationship between instances and ignoring the relationship between ground objects in RSIs\cite{unsupervised_dense,GLCNet}. It obtains features at the instance level, with feature adaptation bias for RSI semantic segmentation tasks that require pixel-level or object-level features\cite{DetCo,InsLoc,PatchReID,xie2021propagate,SoCo,contextual,unsupervised_dense}.

To address the positive sample confounding issue, ContrastiveCrop\cite{ContrastiveCrop} and Leopart\cite{Leopart} use the activation map forwarded from the image to the feature layer to guide sampling, and construct higher quality positive samples. However, these methods ignore the gradient of the contrastive loss backpropagated to the feature layer, and do not make full use of the discriminative information contained in the contrastive loss. Recent works such as LCR\cite{shu2023learning} add an additional branch to align the feature activation map and the contrastive loss gradient activation map, which effectively improves the performance of self-supervised contrastive learning on fine-grained visual recognition. However, this method only changes the feature of the samples, and does not use the activation map to reconstruct the positive samples. For the RSI semantic segmentation task, the positive samples still contain a variety of ground objects, which cannot effectively alleviate the positive sample confounding issue.

To address the feature adaptation bias, DenseCL\cite{DenseCL}, VADeR\cite{unsupervised_dense}, and IndexNet\cite{IndexNet} use dense contrastive approach to optimize the pixel-level contrastive loss between the two views of the input image. However, for the semantic segmentation task of high-resolution remote sensing images, these approaches inevitably lead to higher contrastive learning overhead. In addition, GLCNet\cite{GLCNet} considers adding a local contrastive module for decoder feature maps to the original instance-level contrastive, but this requires the decoder structure for semantic segmentation to be specified in the self-supervised pretraining stage, although the main target of the pretraining stage is to obtain the encoder network.

Unlike the above methods, we observed that the discrimination information contained in the contrastive loss can be mapped to specific regions in RSI through the gradient of unsupervised contrastive loss, these specific regions tend to contain single ground objects. Therefore, we utilize the gradient of the contrastive loss backpropagation to the feature layer to guide sampling and iteratively construct positive and negative samples that contain more singular ground objects during the training process. The major difference between the proposed GraSS and previous work is that the GraSS fully utilizes the contrastive loss gradient to resample the RSI as input to the model, without adding a dense contrastive module or local contrastive module in the pretraining stage. The experimental results indicate that can effectively alleviate the positive sample confounding issue caused by positive samples containing various ground objects, and because the positive and negative samples constructed contain more singular ground objects, our approach will also make the instance-level contrastive closer to the object-level contrastive, effectively mitigating the feature adaptation bias of the instance discrimination pretext task to the downstream semantic segmentation task.

The main contributions of this paper are:
\begin{enumerate}
\item{We propose self-supervised contrastive learning with Gradient guided Sampling Strategy (GraSS) for remote sensing image semantic segmentation, which uses the positive and negative sample discrimination information from the contrastive loss gradient to guide the positive and negative sample construction. It effectively alleviates the positive sample confounding issue and feature adaptation bias of the self-supervised contrastive learning for RSI semantic segmentation, without adding the additional dense contrastive module or local contrastive module.} 
\item{We find that the positive and negative sample discrimination information contained in the contrastive loss gradient can be mapped to specific regions on the RSI, which often contain more singular ground objects. This indicates that the gradient of contrastive loss contains rich feature information, which inspires us to make more use of gradient information to obtain additional model capability in the process of model training.} 
\item{The experimental results on three open datasets, Potsdam, LoveDA Urban, and LoveDA Rural, show that GraSS achieves the best performance compared with eight self-supervised contrastive learning baseline methods from six different types of positive and negative sample construction, and its improved by 1.57\% on average and 3.58\% on maximum of mean intersection over the union (mIoU).}
\end{enumerate}

\section{Related Work}
\subsection{Construction of Positive and Negative Samples}
The construction of positive and negative samples is the basis of self-supervised contrastive learning\cite{SimCLR,cl_hn,RS_SSL1,SSL_Re,UnderstandGV,UnderstandCL}, which usually regards different data augmentations of the same image as positive samples and data augmentations of different images as negative samples\cite{FALSE,SimCLR,MoCo,cl_hn}. The data augmentation method can be divided into two categories according to the different image attributes changed: one is spectral transformation, such as random color distortion\cite{GLCNet}, Gaussian blur\cite{DataAug}, and the other is spatial change, such as random resize crop\cite{DataAug,SimCLR}, random flip\cite{BarlowTwins,DataAug}. Different data augmentation methods have different impacts on the self-supervised contrastive learning model\cite{SimCLR,SimCLRv2,STICL,SeCo,GLCNet}. Among them, the augmentation combination of random resize crop and color distortion has been proven to bring greater performance improvement to the self-supervised contrastive learning model\cite{SimCLR}. In addition, considering the spatio-temporal heterogeneity of RSIs, STICL\cite{STICL} and SeCo\cite{SeCo} make full use of the temporal-shifting characteristics of RSIs to propose a positive and negative sample construction method that is more adaptable to remote sensing image processing.

However, due to the richness, complexity, and imbalance of remote sensing images, the self-supervised contrastive learning model for RSI semantic segmentation suffers from severe sample confounding issue\cite{FALSE,RS_SSL_Re,TOV} : First, negative sample confounding issue, which is often called false negative sample issue\cite{FN_1,FN_2,FALSE}. In order to solve the false negative sample issue, FALSE\cite{FALSE} considers the self-correcting signal based on positive samples and true negative samples giving feedback to the model to guide the model to improve the construction of negative samples and alleviate the false negative sample issue, while IFND\cite{FN_2} and FNC\cite{FN_1} considers the semantic structure of feature space to dynamically detect false negative samples.

The second is the positive sample confounding issue, which undermines the identity assumption of self-supervised contrastive learning\cite{SSL_Re,ID_SSL_Ad,Pre_SSL,UnderstandCL} and is the focus of this paper. In order to solve the positive sample confounding issue, some recent research\cite{ContrastiveCrop,Leopart} used the feature activation maps of images to select specific regions from the original images to generate positive samples. These approach aims to obtain positive samples with semantic consistency guarantees using the activation information forward propagated from the image to the model feature layer. In addition, recent works such as LCR\cite{shu2023learning} introduce the information of contrastive loss gradient, and consider adding a GradCAM fitting branch (GFB)\cite{shu2023learning} to the original contrastive learning model to align the feature activation map and the contrastive loss gradient activation map, which effectively improves the performance of self-supervised contrastive learning on fine-grained visual recognition. However, different from our proposed GraSS, the LCR adds an additional branch and loss function, and only changes the features of the sample, the activation map is not used to guide the resampling. For the RSI semantic segmentation, the positive samples still contain a variety of ground objects, which makes it difficult to effectively alleviate the positive sample confounding issue.

\subsection{Dense Contrastive Learning}
The instance discrimination pretext task of self-supervised contrastive learning acquires image instance-level features, which is naturally adapted to image-level downstream tasks such as RSI scene classification\cite{GLCNet,IndexNet,DetCo,InsLoc,PatchReID,RS_SSL1,RS_SSL_Re}, but suffers from feature adaptation bias for RSI semantic segmentation that requires object-level or pixel-level\cite{DetCo,InsLoc,PatchReID,xie2021propagate,SoCo,unsupervised_dense}. A natural idea for mitigating feature adaptation bias is to optimize the pixel-level contrastive loss between the two views of the input image by dense contrastive approach, giving the model the ability to capture features at the pixel-level or object-level of the image. Methods such as IndexNet\cite{IndexNet}, DenseCL\cite{DenseCL}, and VADeR\cite{unsupervised_dense} add a dense contrastive module to the original instance-level contrastive and obtain stable performance gains in RSI semantic segmentation and object detection, but this inevitably leads to higher computational overhead.

In addition, GLCNet\cite{GLCNet} considers adding a local contrastive module for semantic segmentation decoder feature maps to the original instance-level contrastive, but this requires the decoder structure for semantic segmentation to be specified in the self-supervised pretraining stage.

\begin{figure*}[htp]
\centering
\includegraphics[width=14cm]{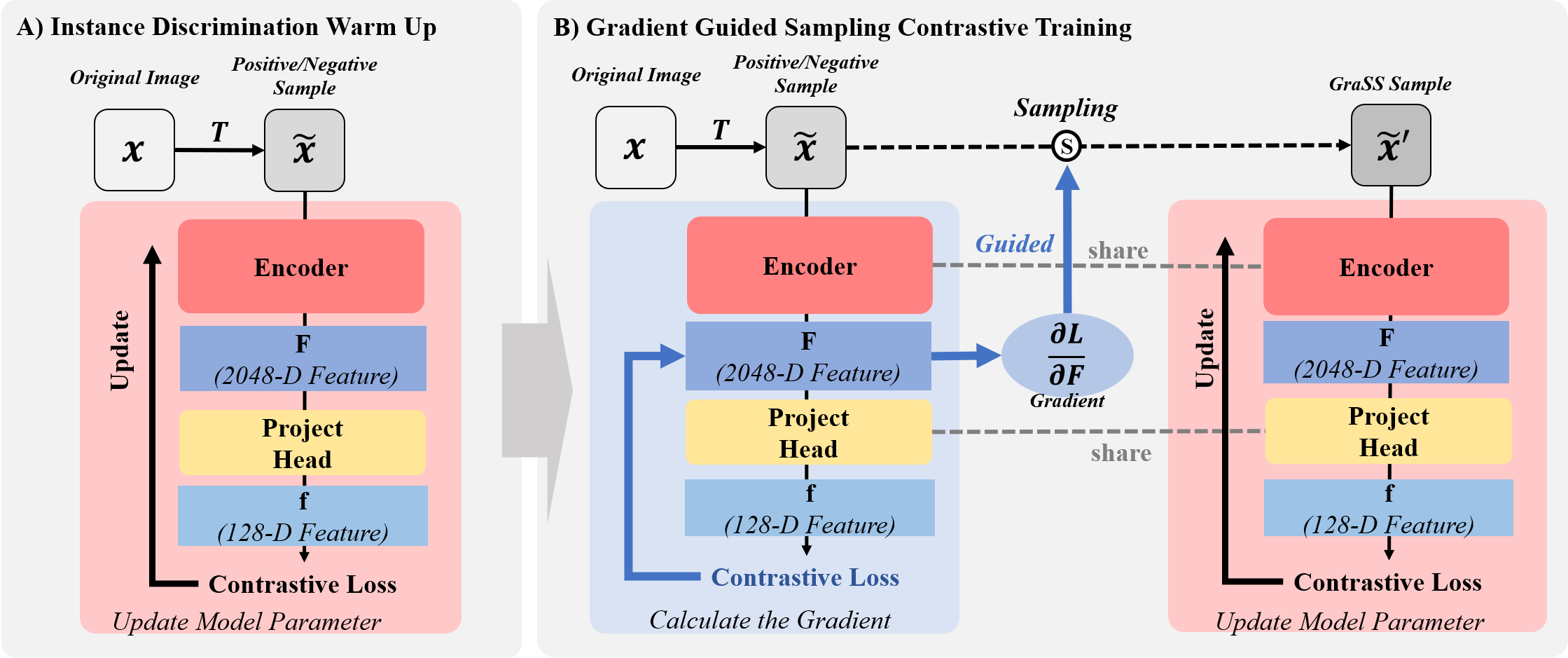} 
\caption{Overview of contrastive learning with Gradient guided sampling strategy (GraSS) for remote sensing image semantic segmentation. A) is the Instance Discrimination Warm Up, the original image is processed by the augmentation function $T$ to obtain positive and negative samples, and the positive and negative samples are input into the model to calculate the contrastive loss and update the model parameters; B) is Gradient Guided Sampling Contrastive Training, the original image is processed by the augmentation function $T$ to obtain positive and negative samples, and the positive and negative samples are input into the model to calculate the gradient, and then the new samples obtained by gradient-guided sampling are input into the model to calculate the contrastive loss and update the model parameters.} 
\label{fig_grass}
\end{figure*}

\section{Method}
\subsection{Overview}

The core idea of the method is derived from the basic characteristics of self-supervised contrastive learning models: self-supervised contrastive learning constrains the model to obtain image features by designing unsupervised instance discrimination pretext task, which can be seen as an image classifier that treats each RSI sample as an independent category. Inspired by the fact that deep network image classifiers tend to rely on a major region of an instance and ignore information about other regions when discriminating between different image instances\cite{cam,wsss,grad_cam}: We expect to obtain the regions of semantic consistency that the self-supervised contrastive learning model focuses on during instance discrimination and use the obtained semantic consistency regions to construct positive and negative samples. We observe that the positive and negative sample discrimination information contained in the contrastive loss gradients can be mapped to specific regions in RSI through the backpropagation of contrasive loss. These specific areas tend to contain single ground objects. Extracting these specific areas as positive and negative samples can effectively solve the positive sample confounding issue and feature adaptation bias of self-supervised contrastive learning for RSI semantic segmentation. 

Therefore, we designed two training stages: 1) instance discrimination warm-up and 2) gradient guided sampling contrastive training. The overall framework of the GraSS is shown in Fig. \ref{fig_grass}. The instance discrimination warm-up stage aims to give the initial positive and negative sample discrimination information to the contrastive loss gradient, which is used to constrain the model to perform instance-level discrimination. The gradient guided sampling contrastive training stage aims to use the gradients of contrastive loss to obtain regions in RSI patches that contain more singular ground objects, in order to construct new positive and negative samples. In this stage, we calculated the contrastive loss twice: the first calculation is to obtain the gradient of the contrastive loss backpropagation to the image feature layer and obtain the activation map. The second calculation is to update the model parameters.

\subsection{Instance Discrimination Warm Up}

The purpose of the instance discrimination warm-up stage is to train the model to acquire initial instance discrimination capabilities, with contrastive loss at this stage used to constrain the model to perform instance discrimination. 
This stage mainly includes the construction of positive and negative sample, model feature extraction, calculation of contrastive loss, and updating of model parameters.

\subsubsection{Construction of Positive and Negative Samples} 
For the RSI data set $x=\{x_i\}^{N}_{i=1}$, it is augmented by function $T$ to get $N \cdot K$ sample instances $\tilde{x}=\{\tilde{x}_{i}\}^{N}_{i=1}$, where $\tilde{x}_i=\{\tilde{x}_{ij}\}^{K}_{j=1}$. The function $T$ consists of three operations: image copy $c(\cdot)$,  random spectral augmentation $rc(\cdot)$, and random spatial augmentation $rs(\cdot)$. Therefore, this process can be described as:

\begin{equation}
    \tilde{x} = T(x) = rs(rc(c(x))) \label{eq: eq1} 
\end{equation}
\begin{equation}
    \tilde{x}^{c}_{i} = c(x_{i}) = [\tilde{x}^{c}_{i1}, \tilde{x}^{c}_{i2}, \dots,\tilde{x}^{c}_{iK}] \label{eq: eq2} 
\end{equation}
\begin{equation}
    \tilde{x}^{Rc}_{i} = rc(x^{c}_{i}) = [\tilde{x}^{Rc}_{i1}, \tilde{x}^{Rc}_{i2}, \dots,\tilde{x}^{Rc}_{iK}] \label{eq: eq3} 
\end{equation}
\begin{equation}
    \tilde{x}_{i} = rs(x^{rc}_{i}) = [\tilde{x}_{i1}, \tilde{x}_{i2}, \dots,\tilde{x}_{iK}] \label{eq: eq4} 
\end{equation}

where $\tilde{x}^{c}_{i1}=\tilde{x}^{c}_{i2}=\dots=\tilde{x}^{c}_{iK}=x_{i}$. All image instances $\tilde{x}_{i}$ within any $\tilde{x}_{ij}$ are obtained from the same original image $x_{i}$,  and they are positive samples of each other. 
Any two image instances $\tilde{x}_{p}$ and $\tilde{x}_{q}$ ($p \not= q$) are negative samples of each other.

\subsubsection{Model Feature Extraction and Contrastive Loss Calculation} 
For image sample instances $\tilde{x}$, we input them into the encoder feature extraction network $E(\cdot)$ to obtain high-dimensional features $\boldsymbol{F}$, and further input high-dimensional features $\boldsymbol{F}$ into the feature projection head $P(\cdot)$ to obtain low-dimensional features $\boldsymbol{f}$ to calculate the contrastive loss $L$, and iteratively update the model parameters. 
Specifically, for the image instance $\tilde{x}_{ij}$, model feature extraction and contrastive loss calculation can be described as:

\begin{equation}
\boldsymbol{F}_{ij} = E(\tilde{x}_{ij}) \label{eq: eq5} 
\end{equation}
\begin{equation}
\boldsymbol{f}_{ij} = P(\boldsymbol{F}_{ij}) \label{eq: eq6} 
\end{equation}
\begin{equation}
l_{ij}= -log(\frac{\sum_{n=1,n\neq j}^{K}{exp(sim(\boldsymbol{f}_{ij},\boldsymbol{f}_{in})/\tau)}}{\sum_{m=1,m\neq i}^{N}\sum_{n=1}^{K}{exp(sim(\boldsymbol{f}_{ij},\boldsymbol{f}_{mn})/\tau)}}) \label{eq: eq7} 
\end{equation}

where $K$ is typically 2, $\tau$ is the temperature parameter, and $sim(\cdot, \cdot)$ usually uses cosine similarity. 
For each iterative parameter update process, the contrastive loss is finally defined as:
\begin{equation}
    L=\frac{1}{N \cdot K} \sum^{N}_{i=1} \sum^{K}_{j=1} l_{ij},
    \label{eq: eq8} 
\end{equation}

\subsection{Gradient Guided Sampling Contrastive Training}

The gradient guided sampling contrastive training stage aims to use the gradient of the contrastive loss to obtain regions in RSI patches that contain more singular ground objects, in order to reconstruct positive and negative samples. This stage involves the construction of positive and negative sample instances, the acquisition of the Discrimination Attention Region (DAR), the reconstruction samples, and the calculation of contrastive loss and model parameter updates. 

\begin{figure}[htp]
\begin{center}
\includegraphics[width=1\linewidth]{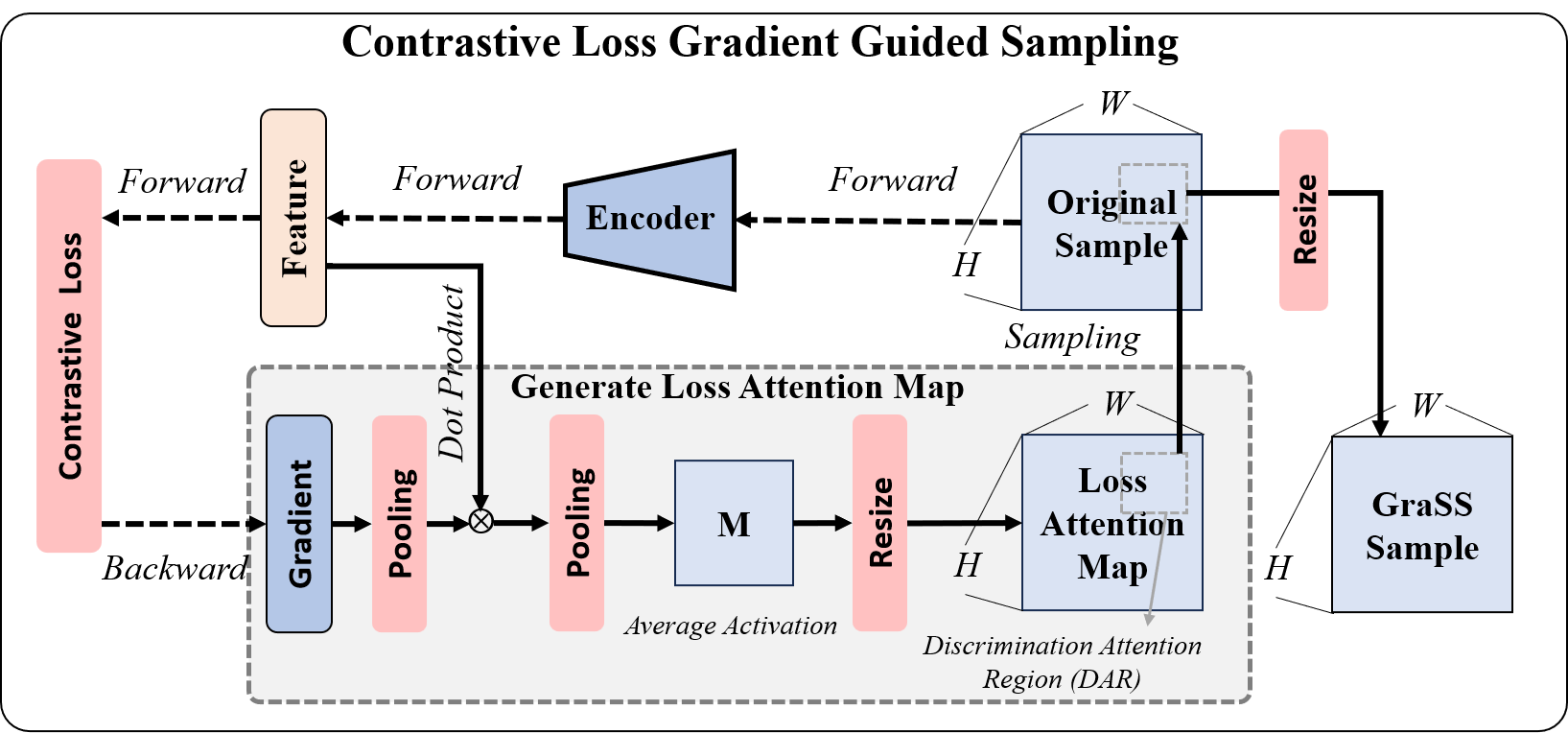}
\setlength{\abovecaptionskip}{0pt} 
\caption{Details of contrastive loss gradient guided sampling. The original samples are input into the model to obtain features and calculate the contrastive loss. Then, the contrastive loss is backpropagated to the feature layer to obtain the gradient to generate the loss attention map. Finally, the original samples were sampled according to the loss attention map to obtain the gradient-guided sampling samples.}
\label{fig_gradientguide}
\end{center}
\end{figure}

The settings for the construction of positive and negative sample instances and calculation of contrastive loss are kept consistent with the instance discrimination warm-up. More details of the acquisition of the Discrimination Attention Region (DAR), the reconstruction samples are shown in Fig. \ref{fig_gradientguide}.

\subsubsection{Acquisition of Discrimination Attention Region (DAR)}
After the construction of positive and negative sample instances using Eq. \eqref{eq: eq1}-\eqref{eq: eq4}, we project the image instances to the low-dimensional features $\boldsymbol{f}$ according to Eq. \eqref{eq: eq5}-\eqref{eq: eq6}, and calculate the contrastive loss using Eq. \eqref{eq: eq7}-\eqref{eq: eq8}) to characterize the distribution of positive and negative sample instances in the feature space. Then, we use backpropagation to calculate the gradient of the contrastive loss to the high-dimensional feature $\frac{\partial L}{\partial \boldsymbol{F}}$. For the image instance $\tilde{x}_{ij}$, this process can be described as:

\begin{equation}
    \frac{\partial L}{\partial \boldsymbol{F}_{ij}} = \frac{\partial L}{\partial l_{ij}} \frac{\partial l_{ij}}{\partial \boldsymbol{f}_{ij}} \frac{\partial \boldsymbol{f}_{ij}}{\partial \boldsymbol{F}_{ij}} \label{eq: eq9} 
\end{equation}

where $\frac{\partial L}{\partial l} = \frac{1}{N \cdot K}$ only related to the amount of original image data $N$ and the number of copies $K$ of the image copy function $c(\cdot)$. 
Then, we calculate the average activation $M$ for the dot product of contrastive loss gradient and the feature $\boldsymbol{F}$ and resize $M$ to the same size as the image sample to obtain the contrastive Loss Attention Map (LAM). For the image instance $\tilde{x}_{ij}$, this process can be described as:
\begin{equation}
    M = \frac{1}{D} \sum^{D}_{d=1} pooling(\frac{\partial L}{\partial \boldsymbol{F}_{ij}})\boldsymbol{F}^{d}_{ij},
    \label{eq: eq10} 
\end{equation}
\begin{equation}
    LAM = \underset{H=h,W=w}{Resize}(M).
    \label{eq: eq11} 
\end{equation}

In the Eq. \eqref{eq: eq10}, $\boldsymbol{F}^{d}_{ij}$ denotes the d-th dimensional component of the D-dimensional feature $\boldsymbol{F}_{ij}$. $pooling(\cdot)$ denotes the global pooling operation applied to the gradients. $\underset{H=h,W=w}{Resize}(\cdot)$ represents the resizing of the activation $M$ into a two-dimensional activation map of height $h$ and width $w$, where $h$ and $w$ correspond to the height and width of the input image instances.

Finally, we obtain the discriminative attention region (DAR) based on the contrastive Loss Attention Map (LAM). Specifically, we define the Discrimination Attention Region acquisition function G(LAM; $T_A$), where $T_A$ is the activation map threshold for selecting DAR. 
We regard the regions in LAM with a value higher than $T_A$ as the candidate discrimination attention region $R$, calculate the maximum activation value of all candidate discrimination attention regions $R$, and select the candidate region with the highest maximum activation value as the Discrimination Attention Region (DAR).
The above process can be described as:
\begin{equation}
    DAR = G(LAM; T_{A} = t),
    \label{eq: eq12} 
\end{equation}
\begin{equation}
    R = \{R_{i}\} = (LAM > t),
    \label{eq: eq13} 
\end{equation}
\begin{equation}
    max(DAR) = max(max(R_{i})),
    \label{eq: eq14} 
\end{equation}

where $R_i$ refers to the $i$-th 4-connected closed region in $R$. 
The rules for the operation of the function $G$, as described in Eq. \eqref{eq: eq12}, are defined by Eq. \eqref{eq: eq13}, and Eq. \eqref{eq: eq14}.

\subsubsection{Reconstruction of Positive and Negative Samples}
We reconstructed positive and negative samples based on the DAR. Specifically, we first obtain the coordinates of the centroid $(x,y)$, width $w$, and height $h$ of the smallest outer rectangle of the DAR corresponding to the original image sample. Afterward, we crop the corresponding RSI region based on the coordinates and resize it to the original image size to obtain a new sample. We refer to the operation of cropping an image based on the DAR as DACrop. The above process can be described as:

\begin{equation}
    x, y, h, w = Box(DAR_{ij}),
    \label{eq: eq15} 
\end{equation}
\begin{equation}
    \tilde{x}_{ij}^{'} = \underset{X=x, Y=y, H=h,W=w}{DACrop}(\tilde{x}_{ij}).
    \label{eq: eq16} 
\end{equation}

Finally, we input the updated image instance $\tilde{x}_{ij}$ into the model to extract features, calculate the contrastive loss and update the model parameters.

The proposed GraSS can be described in Algorithm \ref{alg_grass}.

\begin{algorithm}
	\caption{Pseudocode for GraSS} 
	\label{alg_grass}
	\begin{algorithmic}[1]
		\REQUIRE RSI dataset $X$; encoder E(·), project head P(·); augmentation fucntion T(·); batch size $N$; warm-up epoch $e$, threshold $t$, current model training epoch $e_c$
        \FOR{batch $x$ from $X$}
            \STATE Data Augmentation: $\tilde{x}=T(x)$
            \STATE Get features: $\boldsymbol{f}_{ij} = P(E(\tilde{x}))$
            \STATE Calculate contrastive loss $L$ using Equation \eqref{eq: eq7} and \eqref{eq: eq8}
            \IF{$e_c \leq e$}
                \STATE Update E(·) and P(·)
            \ELSE
                \STATE Get LAM using Equation \eqref{eq: eq9}, \eqref{eq: eq10}, and \eqref{eq: eq11}
                \STATE Get DAR: $DAR = G(LAM; T_{A} = t)$
                \STATE Construct sample $\tilde{x}_{ij}^{'}$ with DACrop using Equation \eqref{eq: eq15} and \eqref{eq: eq16}
                \STATE Get features: $\boldsymbol{f}_{ij}^{'} = P(E(\tilde{x}_{ij}^{'}))$
                \STATE Calculate contrastive loss $L$ using Equation \eqref{eq: eq7} and \eqref{eq: eq8}
                \STATE Update E(·) and P(·)
            \ENDIF 
        \ENDFOR
	\end{algorithmic} 
\end{algorithm}

\section{Experiment}
\subsection{Experimental Setup}

\subsubsection{Dataset}

We selected three high-resolution RSI semantic segmentation datasets Potsdam\cite{isprs_potsdam}, LoveDA Urban, and LoveDA Rural\cite{LovaDA} to evaluate the semantic segmentation performance of the self-supervised contrastive learning model on high-resolution RSIs. TABLE \ref{table_data} shows more detailed information about the Potsdam, LoveDA Urban, and LoveDA Rural datasets.

\begin{table*}[h]
\renewcommand\arraystretch{1.5}
\centering
\caption{Detail information of Potsdam, LoveDA Urban, and LoveDA Rural dataset.}
\begin{tabular}{cccc}
\hline
\textbf{Dataset}                                          & Potsdam                      & LoveDA Urban                 & LoveDA Rural                 \\ \hline
Resolution (m)                                            & 0.05                         & 0.3                          & 0.3                          \\
Crop Size                                                 & 256$\times$ 256              & 256$\times$ 256              & 256$\times$ 256              \\
The amount of data for self-supervised pretraining        & 13824                        & 18496                        & 21856                        \\
The amount of data for semantic segmentation fine-tuning  & 138                          & 184                          & 218                          \\
The amount of data for semantic segmentation test dataset & 8064                         & 10832                        & 15872                        \\ \hline
\end{tabular}
\label{table_data}
\end{table*}

\subsubsection{Baselines}
We selected eight state-of-the-art methods from six different types of positive and negative sample construction as baselines to evaluate the performance of GraSS.
\paragraph{Original Contrastive Learning Method}
We selected the representative SimCLR\cite{SimCLR} and MoCo v2\cite{MoCov2} as the classical self-supervised contrastive learning baselines. Both SimCLR\cite{SimCLR} and MoCo v2\cite{MoCov2} use the typical positive and negative sample construction methods, where different augmentations of the same image are treated as positive samples and augmentations of different images are treated as negative samples. The difference is that SimCLR constructs negative samples from other images in the same training batch, which limits the number of negative samples by the batch size\cite{SimCLR}, while MoCo v2 updates the negative samples by maintaining a queue momentum, and the number of negative samples is not limited by the number of samples in the training batch\cite{MoCov2,MoCo}.

\paragraph{Contrastive Learning Method with Clustering}
We selected PCL\cite{PCL} as a self-supervised contrastive learning baseline that introduces clustering to build positive and negative samples. 
Unlike SimCLR and MoCo v2, PCL introduces a clustering strategy to construct positive and negative samples based on data augmentation, which treats cluster centers of the same class of image clustering as positive samples and cluster centers of different classes of clusters as negative samples\cite{PCL}.

\paragraph{Contrastive Learning Method without Negative Samples}
We selected Barlow Twins\cite{BarlowTwins} and BYOL\cite{BYOL} as self-supervised contrastive learning baselines that do not construct negative samples and only construct positive samples. To avoid the model collapse caused by only bringing positive samples closer, BYOL uses an asymmetric network structure to project different augmentations of the same image into different feature spaces for comparison, while Barlow Twins does not directly pull positive samples closer, it only constrains the dimensions of sample features to be relatively independent.

\paragraph{Negative Aware Contrastive Learning Method}
We selected FALSE\cite{FALSE} as a self-supervised negative aware contrastive learning baseline. FALSE adds a determination module to correct false negative samples into positive samples based on the classical positive and negative sample construction method, the positive samples are not only derived from different augmentations of the same image but also from images that contain the same ground objects as the positive samples\cite{FALSE}.

\paragraph{Positive Aware Contrastive Learning Method}
We selected ContrastiveCrop\cite{ContrastiveCrop} as a self-supervised positive aware contrastive learning baseline. ContrastiveCrop uses the activation information propagated forward from the image to the feature layer of the model to construct positive samples\cite{ContrastiveCrop}.

\paragraph{Dense Contrastive Learning Method}
We selected DenseCL\cite{DenseCL} as a self-supervised contrastive learning baseline with the dense contrastive module. DenseCL adds dense feature contrastive constraints to the instance contrastive. It gives the model the ability to capture certain object-level or pixel-level features of the image\cite{DenseCL}.

\subsubsection{Metrics}
We selected three metrics, the mean Intersection-over-Union (mIoU), the Overall Accuracy (OA), and the mean class Accuracy (mAcc) to quantitatively evaluate the performance of the self-supervised contrastive model on the test dataset for the downstream semantic segmentation task. The mIoU is a common metric for the semantic segmentation task, and for a single ground object, the intersection-over-Union (IoU) is defined by the following equation:

\begin{equation}
IoU = \frac{prediction \cap target}{prediction \cup target} \label{eq: eq17} 
\end{equation}

Where prediction refers to the predicted result of the model for the ground object, and target refers to the ground truth of the ground object. The mIoU is equal to the average of the IoU of all objects. 

The OA represents the overall accuracy of the predicted result on the test dataset, which is defined by the following equation:

\begin{equation}
OA = \frac{TP}{N} \label{eq:eq18} 
\end{equation}

where the TP means the total number of pixels that are correctly predicted, and the N means the total number of pixels.

Slightly different from OA, mAcc is used to indicate the average level of accuracy of the predicted result for each ground object class. Specifically, the prediction Accuracy (Acc) for a single ground object class can be defined by the following equation:

\begin{equation}
Acc = \frac{{TP}_i}{N_i} \label{eq:eq19}
\end{equation}

where the $TP_{i}$ means the number of correctly predicted pixels for a specific ground object class and the $N_{i}$ means the number of pixels for a specific ground object class in ground truth. The mAcc is the average of the Acc of all ground object classes.

\subsubsection{Implementation Details}
For both the eight self-supervised contrastive learning baselines and the proposed GraSS, we used ResNet50\cite{ResNet} as the backbone network. 

In the self-supervised pretraining stage, we train 350 epochs using the entire training dataset without labels, and the batch size is set to 256. For each baseline method, we use the data augmentation and optimization settings recommended in the original paper, and all self-supervised pretraining methods were used to train the feature extractor only. For the proposed GraSS, the number of instance discrimination warm-up training epochs is also included in the total number of pretraining epochs for a fair comparison with baselines.

In the RSI semantic segmentation fine-tuning stage, in order to accurately evaluate the performance of the features extracted from different self-supervised pretraining methods, we freeze the weights of the entire backbone network and update only the parameters of the feature decoder used to obtain the RSI semantic segmentation results. We randomly select 1\% of the entire training dataset for fine-tuning training. For the eight baseline methods and proposed GraSS, the randomly selected fine-tuning data were kept consistent. We uniformly use the Stochastic Gradient Descent (SGD) optimizer\cite{SGD} for fine-tuning training for 150 epochs with the batch size set to 16.

\subsection{Experimental Result}

\begin{table*}[htp]
\renewcommand\arraystretch{1.5}
\caption{Quantitative comparison results with eight state-of-the-art self-supervised contrastive learning baseline methods and GLCNet.\label{PA1}}
\centering
\setlength{\tabcolsep}{2mm}{
\begin{threeparttable}
    \begin{tabular}{cccccccccccc}
    \hline
    \multirow{2}{*}{Method} &\multirow{2}{*}{Pretraining Module} & \multicolumn{3}{c}{Potsdam}                      & \multicolumn{3}{c}{LoveDA Urban}                 & \multicolumn{3}{c}{LoveDA Rural}  & \multirow{2}{*}{\makecell[c]{Time\tnote{*}\\(Training/Test)}} \\ \cline{3-11} 
                            &                 & OA             & mIoU           & mAcc           & OA             & mIoU           & mAcc           & OA             & mIoU           & mAcc           \\ \hline
    SimCLR                  & Encoder         & 61.18          & 43.02          & 54.90          & 41.91          & 33.09          & 45.31          & {\ul 62.96}    & {\ul 41.30}    & {\ul 52.97}    & 3.0h/12min\\
    MoCo v2                 & Encoder         & 60.21          & 42.81          & 54.53          & 40.61          & 32.92          & 45.77          & 58.01          & 36.28          & 47.84          & 2.5h/12min\\
    PCL                     & Encoder         & 61.45          & 43.13          & 55.15          & 40.07          & 33.28          & 45.99          & 59.40          & 37.42          & 49.41          & 8.5h/12min\\
    Barlow Twins            & Encoder         & 61.42          & 43.17          & 54.95          & {\ul 43.05}    & {\ul 34.32}    & {\ul 46.09}    & 56.29          & 33.93          & 51.55          & 7.5h/12min\\
    BYOL                    & Encoder         & {\ul 61.54}    & 43.93          & 55.73          & 35.18          & 28.49          & 39.94          & 60.21          & 37.39          & 48.34          & 5.0h/12min\\
    FALSE                   & Encoder         & 60.65          & 43.12          & 55.45          & 42.44          & 33.69          & 46.08          & 62.44          & 40.91          & 51.82          & 3.5h/12min\\
    ContrastiveCrop         & Encoder         & 60.89          & 42.86    & 54.64    & 40.24          & 33.35          & 45.33          & 61.98          & 39.28          & 51.85          & 7.5h/12min              \\ 
    DenseCL                 & Encoder         & 61.44          & {\ul 44.34}    & {\ul 56.40}    & 37.01          & 30.85          & 42.07          & 62.85          & 38.69          & 49.62          & 5.5h/12min\\ 
    \textbf{GraSS(Ours)}    & \textbf{Encoder}         & \textbf{62.28} & \textbf{44.39} & \textbf{56.50} & \textbf{43.68} & \textbf{34.77} & \textbf{46.77} & \textbf{65.25} & \textbf{42.58} & \textbf{53.79} & 4.5h/12min\\ \hline
    GLCNet                  & Encoder+Decoder & 77.91          & 60.57          & 75.26          & 50.78          & 41.26          & 56.55          & 63.36          & 40.41          & 53.36 & 8.5h/12min\\ 
    \textbf{GraSS(Ours)}    & \textbf{Encoder+Decoder} & \textbf{78.79} & \textbf{61.41} & \textbf{75.62} & \textbf{52.60} & \textbf{42.05} & \textbf{57.82} & \textbf{65.65} & \textbf{40.96} & \textbf{57.62} & 9.5h/12min\\ \hline
    \end{tabular}
    \begin{tablenotes}
        \footnotesize
        \item[*] The training time is measured on the Potsdam training dataset using an A800 GPU for 350 epochs, and the test time is measured on the Potsdam test dataset using an RTX 3090 GPU for 150 epochs.
    \end{tablenotes}  
\end{threeparttable}
}
\end{table*}
\unskip

\begin{table}[htp]
\renewcommand\arraystretch{1.5}
\caption{The semantic segmentation results of GLCNet with gradient guided sampling strategy.\label{PA2}}
\centering
\setlength{\tabcolsep}{1.4mm}{\begin{tabular}{cccc}
\hline
\multicolumn{2}{c}{\multirow{2}{*}{Dataset and Metric}}   & \multicolumn{2}{c}{GLCNet}                                  \\ \cline{3-4} 
                         &                     & \makecell[c]{w/o\\Gradient Guided}    & \makecell[c]{\textbf{w/}\\\textbf{Gradient Guided}}  \\ \hline
\multirow{4}{*}{Potsdam}        &Kappa & 71.77 & \textbf{72.79} \\
                                &OA    & 77.91 & \textbf{78.79} \\
                                &mIoU  & 60.57 & \textbf{61.41} \\
                                &mAcc  & 75.26 & \textbf{75.62} \\ \hline
\multirow{4}{*}{LoveDA Urban}   &Kappa & 42.08 & \textbf{43.23} \\
                                &OA    & 50.78 & \textbf{52.60} \\
                                &mIoU  & 41.26 & \textbf{42.05} \\
                                &mAcc  & 56.55 & \textbf{57.82} \\ \hline
\multirow{4}{*}{LoveDA Rural}   &Kappa & 49.22 & \textbf{50.45} \\
                                &OA    & 63.36 & \textbf{65.65} \\
                                &mIoU  & 40.41 & \textbf{40.96} \\
                                &mAcc  & 53.36 & \textbf{57.62} \\   
\hline 
\end{tabular}}
\end{table}

\begin{table}[htp]
\renewcommand\arraystretch{1.5}
\caption{Comparison of semantic segmentation results between GraSS and two state-of-the-art sampling method.\label{PA_samplem}}
\centering
\setlength{\tabcolsep}{1.4mm}{\begin{tabular}{cccccc}
\hline
\multicolumn{2}{c}{\multirow{2}{*}{Dataset and Metric}}   & \multicolumn{4}{c}{Sampling Method} \\ \cline{3-6}
                                &      & Original  & ContrastiveCrop  & LCR     & \textbf{Ours} \\ \hline
\multirow{3}{*}{Potsdam}        &OA    & 60.21     & 60.89            & 61.26   & \textbf{62.28} \\
                                &mIoU  & 42.81     & 42.86            & 43.88   & \textbf{44.39} \\
                                &mAcc  & 54.53     & 54.64            & 55.52   & \textbf{56.50} \\ \hline
\multirow{3}{*}{LoveDA Urban}   &OA    & 40.61     & 40.24            & 42.50   & \textbf{43.68} \\
                                &mIoU  & 32.92     & 33.35            & 33.53   & \textbf{34.77} \\
                                &mAcc  & 45.77     & 45.33            & 45.97   & \textbf{46.77} \\ \hline
\multirow{3}{*}{LoveDA Rural}   &OA    & 58.01     & 61.98            & 64.85   & \textbf{65.25} \\
                                &mIoU  & 36.28     & 39.28            & 40.39   & \textbf{42.58} \\
                                &mAcc  & 47.84     & 51.85            & 50.26   & \textbf{53.79} \\  
\hline 
\end{tabular}}
\end{table}

\begin{figure*}[htp]
\vspace{-5mm}
\centering
\includegraphics[width=17cm]{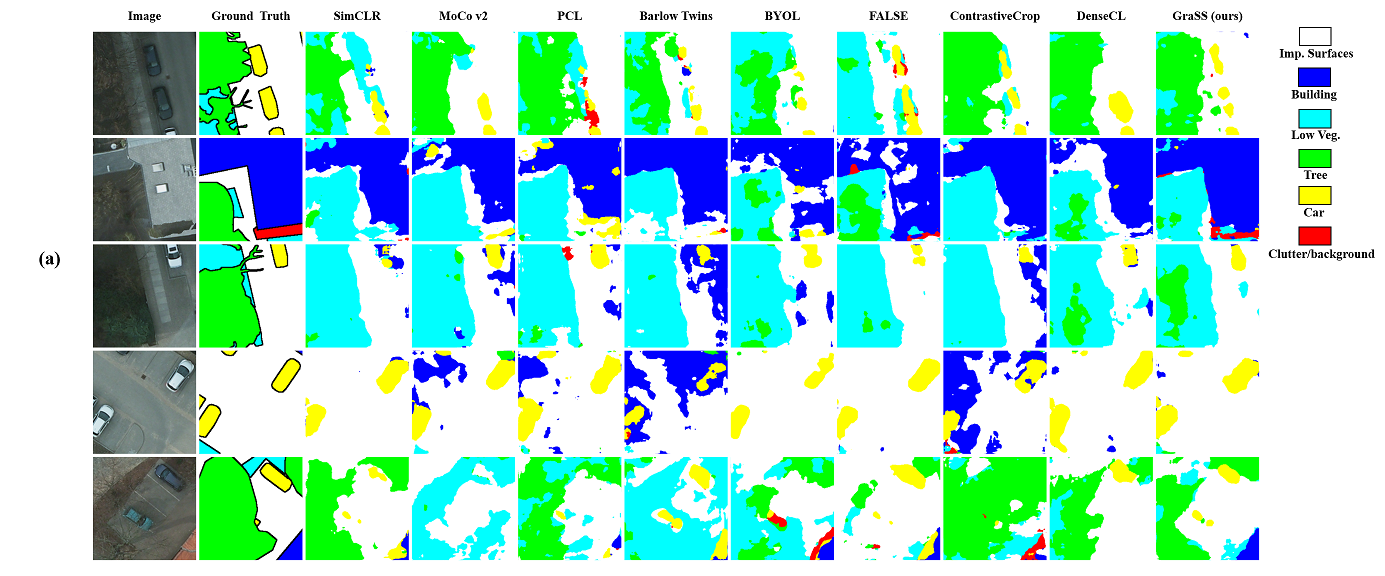}
\includegraphics[width=17cm]{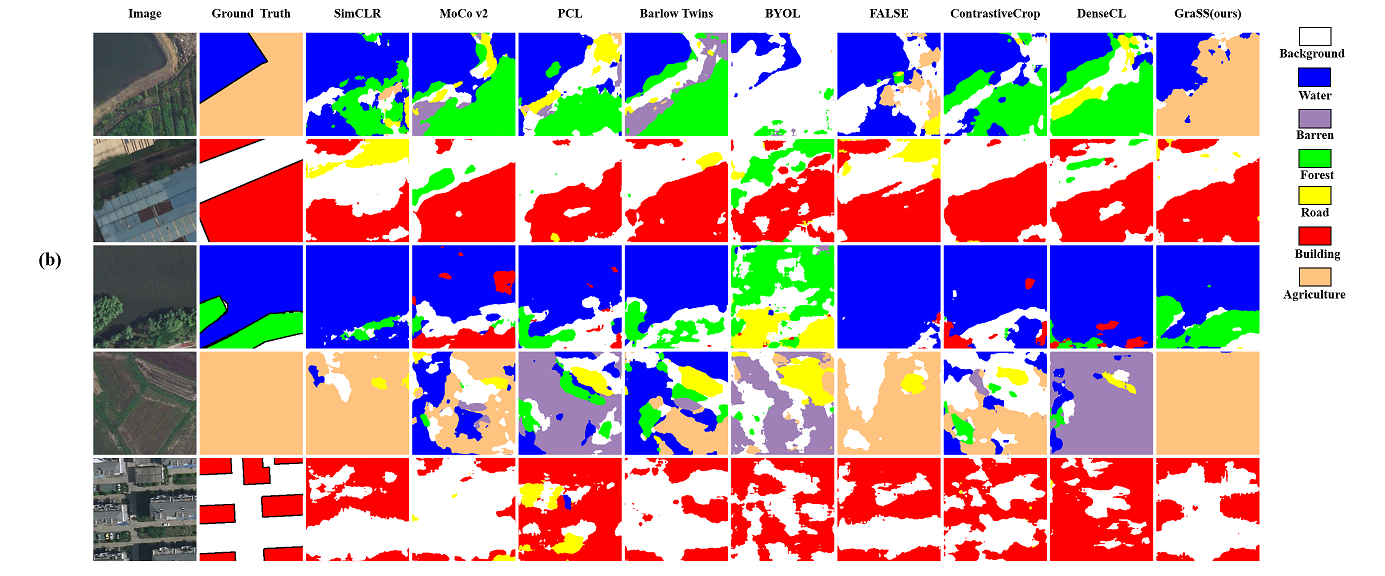} 
\includegraphics[width=17cm]{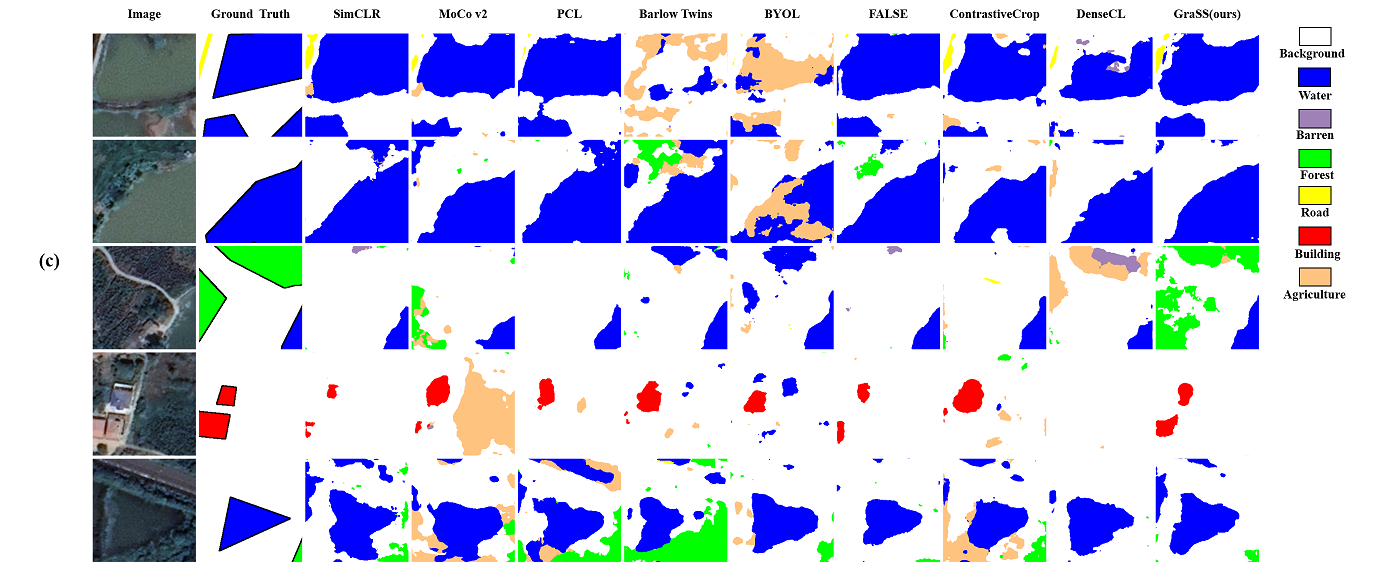}
\caption{Qualitative comparison results with eight self-supervised contrastive learning baseline methods. Where the five RSIs in a) are from the test dataset of Potsdam, the five RSIs in b) are from the test dataset of LoveDA Urban, and the five RSIs in c) are from the test dataset of LoveDA Rural.} 
\label{fig_pa}
\end{figure*}

In this section, we present five aspects of evaluation for the proposed GraSS.

First is performance analysis, we compare the proposed GraSS with six types, a total of eight self-supervised contrastive learning baseline methods on the Potsdam, LoveDA Urban, and LoveDA Rural datasets, and provide both quantitative and qualitative analysis results. In order to explore the applicability of the gradient guided sampling strategy, unlike the eight baseline methods mentioned above that train only the feature extractor in the pre-training stage, we also compare with GLCNet\cite{GLCNet}, which requires a specified RSI semantic segmentation decoder in the pre-training stage. In addition, to further validate the performance of the proposed GraSS, we also compare GraSS with two sampling methods, ContrastiveCrop\cite{ContrastiveCrop} and LCR\cite{shu2023learning}.

Second is the ablation study, we perform ablation experiments to verify the effectiveness of each module of the proposed GraSS.

Third is performance sensitivity analysis, we examine the effects of two hyperparameters corresponding to the two training stages of GraSS: the instance discrimination warm-up epoch and the threshold $T_A$ of the activation map for selecting DAR on the RSI semantic segmentation performance of the model. The instance discrimination warm-up epoch hyperparameter corresponds to the instance discrimination warm-up stage, and the threshold of activation map hyperparameter corresponds to the gradient guided sampling contrastive training stage.

Fourth is the analysis of the number of ground objects contained in the sample. We quantitatively evaluated the number of ground objects contained in the samples and found that positive and negative samples obtained by GraSS contain more singular ground object types compared to the samples obtained by original self-supervised contrastive methods.

Finally is the visual analysis of the contrastive Loss Attention Map (LAM). We examined the effect of the warm-up epoch and the ground objects contained in RSI on the LAM.

\subsubsection{Performance analysis}

\begin{figure*}[htp]
\vspace{-5mm}
\begin{center}
\includegraphics[width=1\linewidth]{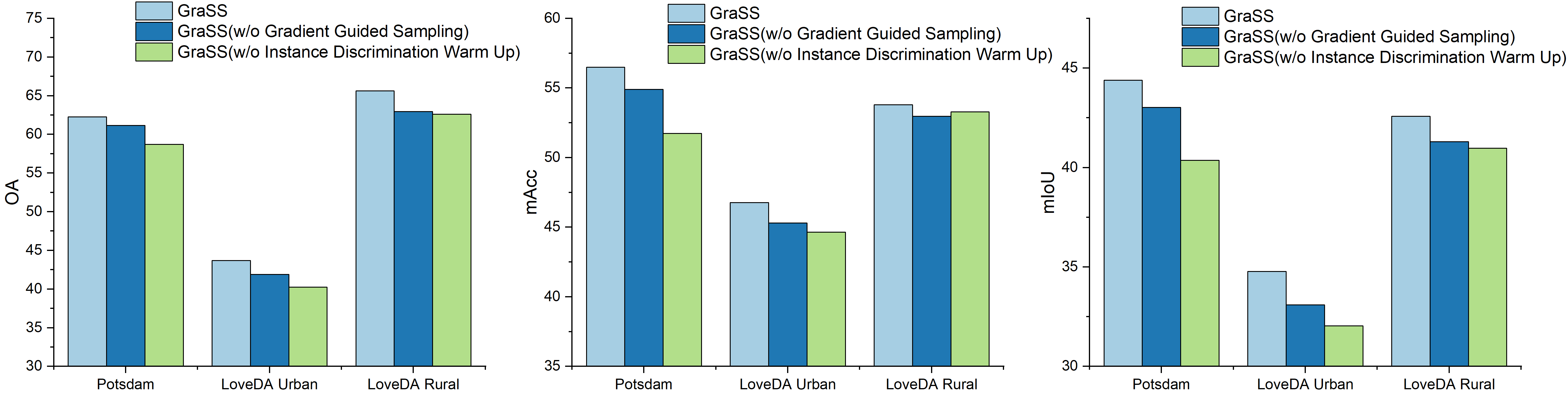}
\setlength{\abovecaptionskip}{0pt} 
\caption{Results of the ablation study exploring the effectiveness of each module of the proposed GraSS.}
\label{fig_abs}
\end{center}
\end{figure*}

\paragraph{Quantitative Analysis}

To evaluate the performance of the proposed GraSS, we first compared it with six types, a total of eight self-supervised contrastive learning baseline methods and GLCNet\cite{GLCNet}, which requires a specified RSI semantic segmentation decoder in the pre-training stage, and the experiment results are shown in TABLE \ref{PA1}. In addition, in order to explore the applicability of the gradient guided sampling strategy, we align the experimental conditions and metrics of GraSS with GLCNet to show the performance improvement of GLCNet with the gradient guided strategy, and the experimental results are shown in TABLE \ref{PA2}. Finally, in order to further verify the performance of the proposed GraSS, we also compared GraSS with the two sampling methods: ContrastiveCrop\cite{ContrastiveCrop} and Learning Common Rationale (LCR)\cite{shu2023learning}, and the experimental results are shown in TABLE \ref{PA_samplem}.

TABLE \ref{PA1} shows that GraSS achieves the best results on all three metrics for the three datasets compared to the eight self-supervised contrastive learning baseline methods of SimCLR, MoCo v2, PCL, Barlow Twins, BYOL, FALSE, ContrastiveCrop, and DenseCL. On the Potsdam dataset, GraSS performs only slightly outperformed DenseCL, but on the LoveDA Urban dataset, GraSS is 3.85\% higher than DenseCL in terms of mIoU, and on the LoveDA Rural dataset, GraSS is 3.89\% than DenseCL in terms of mIoU, this demonstrates the stable semantic segmentation performance improvement of GraSS.

In addition, compared to the original self-supervised contrastive learning method SimCLR, we observed that MoCo v2, PCL, Barlow Twins, BYOL, and DenseCL show significant performance degradation on the LoveDA Rural dataset, with DenseCL and BYOL also showing significant performance degradation on the LoveDA Urban dataset, exhibiting unstable semantic segmentation performance.

TABLE \ref{PA2} shows the experimental results of using the gradient guided sampling strategy for the GLCNet that requires a specified semantic segmentation decoder in the self-supervised pretraining stage. The experimental results indicate that the gradient guided sampling strategy further improves the semantic segmentation performance of the GLCNet. Meanwhile, It indicates that the gradient guided sampling strategy is also applicable to the self-supervised contrastive learning method that trains both the feature extractor and the semantic segmentation decoder in the pretraining stage.

TABLE \ref{PA_samplem} shows the comparison results of semantic segmentation performance of the proposed GraSS with Original, ContrastiveCrop, and LCR, where Original represents the original self-supervised contrastive learning method. The experimental results indicate that on Potsdam, LoveDA Urban, and LoveDA Rural datasets, compared with Original, ContrastiveCrop, and LCR, the GraSS achieves the best performance on three indicators. Specifically, on the LoveDA Rural dataset, the proposed GraSS improves the mIoU by 2.19\% and the mAcc by 3.53\% compared with the best baseline.

\paragraph{Qualitative Analysis}

We qualitatively analyzed the visualization results of semantic segmentation of RSIs, and the experimental results are shown in Fig. \ref{fig_pa}. The experimental results show that the semantic segmentation results of GraSS present richer details than the eight self-supervised contrastive learning baselines, especially for small-scale ground objects in the RSIs, which are difficult to be captured by the instance-level self-supervised contrastive learning methods. In addition, the GraSS also presents a more stable semantic segmentation effect for RSIs containing a large range of homogeneous ground objects.

Quantitative and qualitative experimental results indicate that the proposed GraSS effectively improves the performance on RSI semantic segmentation tasks, and performs better in both quantitative and qualitative aspects. This is because the proposed GraSS fully utilizes the discriminative information in the contrastive loss gradient to construct samples containing more singular ground objects, which effectively alleviates the positive sample confounding issue in the process of contrastive learning. Meanwhile, the model can benefit from the contrastive of samples containing a single ground object, and obtain more accurate features of ground objects.

Although the proposed GraSS effectively alleviates the positive sample confounding issue of self-supervised contrastive learning in RSI semantic segmentation tasks compared with other methods, since the process of constructing positive and negative samples is unsupervised, the proposed GraSS cannot absolutely guarantee that the obtained samples only contain a single type of ground object, and cannot completely eliminate the positive sample confounding issue.

\subsubsection{Ablation study}
In order to explore the effectiveness of each module of the proposed GraSS, we conducted an ablation study on three datasets: Potsdam, LoveDA Urban, and LoveDA Rural. Three sets of experiments were conducted, the first set of experiments used the complete GraSS, the second set of experiments removed the gradient guided sampling module on the basis of the proposed GraSS, and the third set of experiments removed the instance discriminative warm-up training module on the basis of the GraSS. The experimental results are shown in Fig. \ref{fig_abs}.

The experimental results in Fig. \ref{fig_abs} show that each module of GraSS improves the model performance. Compared with complete GraSS and the GraSS without gradient guided sampling, GraSS without instance discrimination warm-up demonstrates the worst performance on the above three datasets. This indicates that it is necessary to conduct instance discrimination warm up in the initial stage of the model training and train the model to obtain the initial instance discrimination ability. The instance discrimination warm up gives discriminative information to the contrastive loss gradient, which can help the gradient-guided sampling to further improve the performance of the model.

In addition, compared with the complete GraSS, the performance of the GraSS without gradient guided sampling decreases in all indicators of the three datasets, which indicates that the gradient guided sampling contrastive training can effectively improve the semantic segmentation performance of the model after the instance discrimination warm up.

\begin{figure*}[htp]
\centering
\includegraphics[width=0.9\textwidth]{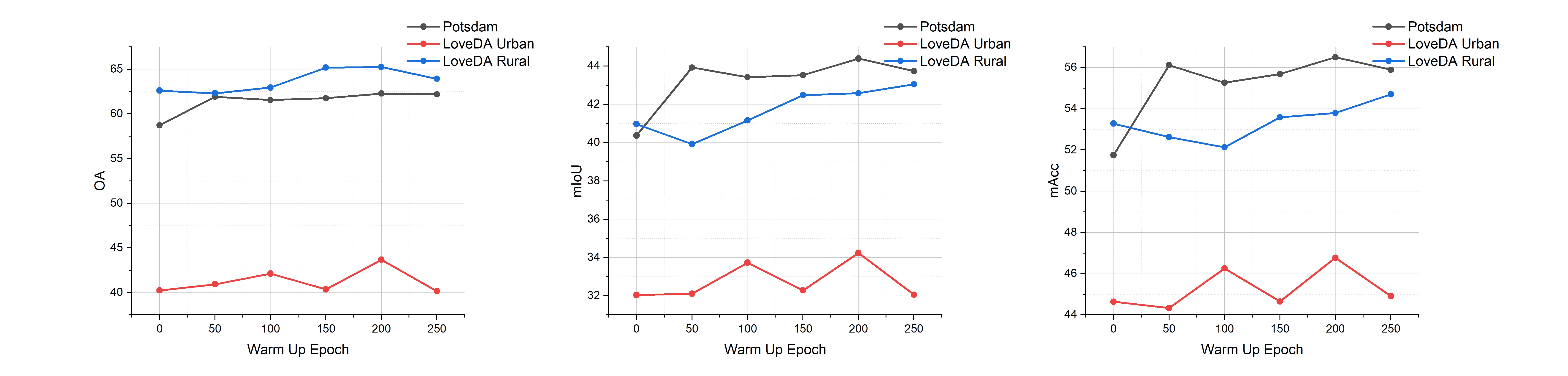}
\setlength{\abovecaptionskip}{0pt} 
\caption{Semantic segmentation results with 6 different instance discriminationwarm-up epochs.}
\label{fig_wa}
\end{figure*}

\begin{figure*}[htp]
\centering
\includegraphics[width=0.9\textwidth]{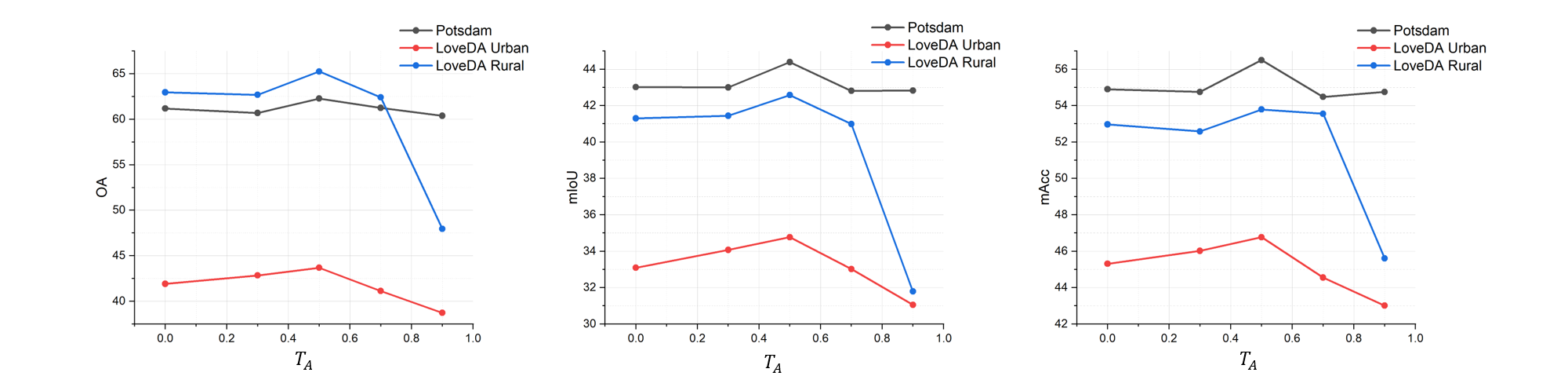}
\setlength{\abovecaptionskip}{0pt} 
\caption{Semantic segmentation results with 5 different threshold $T_A$ of the activation map for selecting DAR.}
\label{fig_ta}
\end{figure*}

\begin{figure}[ht]
\vspace{-5mm}
\begin{center}
\includegraphics[width=1\linewidth]{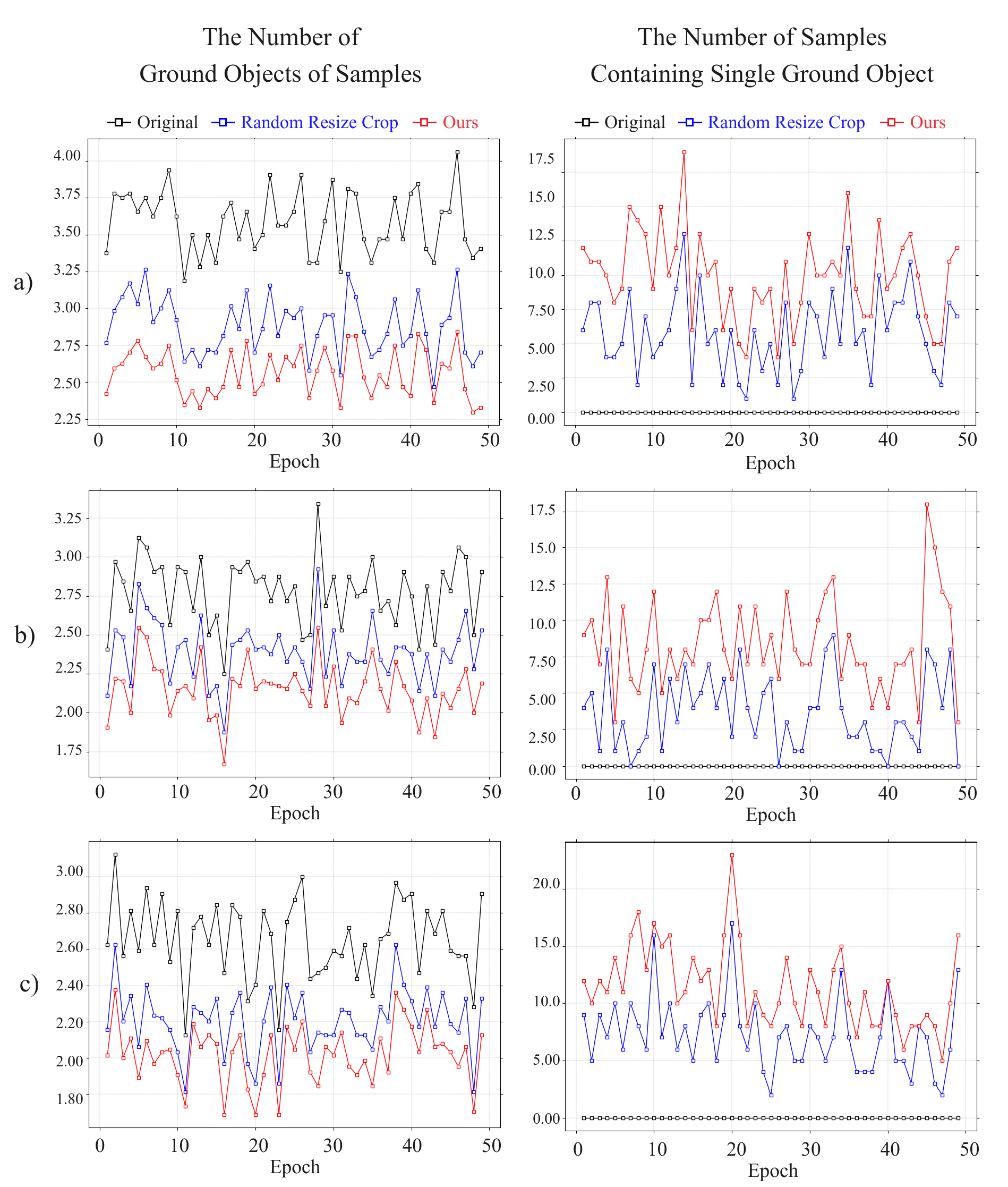}
\setlength{\abovecaptionskip}{0pt} 
\caption{Result of analysis of the Number of Ground Objects Contained in the Sample. The first-row a) shows the results on the Potsdam dataset, the second-row b) shows the results on the LoveDA Urban dataset, and the third-row c) shows the results on the LoveDA Rural dataset.}
\label{fig_num}
\end{center}
\end{figure}

\subsubsection{Performance sensitivity analysis}
In this section, we examine the effects of two hyperparameters of GraSS: the instance discrimination warm-up epochs and the threshold of the activation map for selecting DAR on the performance of semantic segmentation. 
The instance discrimination warm-up epochs hyperparameter corresponds to the first training stage of GraSS and the threshold of the activation map for selecting the DAR hyperparameter corresponds to the second training stage of GraSS.

\paragraph{Analysis of the Instance Discrimination Warm-Up Epochs}

The self-supervised contrastive learning methods with gradient guided sampling rely on the instance discrimination ability of the model. Therefore, we analyze the impact of the warm-up training epochs on the semantic segmentation performance of the GraSS.
We fixed the gradient guided sampling training epochs to 150, and selected six instance discrimination warm-up epochs of 0, 50, 100, 150, and 200 on the Potsdam, LoveDA Urban, and LoveDA Rural datasets for comparison. 

Fig. \ref{fig_wa} shows the semantic segmentation results for the GraSS with six different warm-up epochs. The experimental results show that for the Potsdam, LoveDA Rural, and LoveDA Urban datasets, when the instance discrimination warm-up epoch is increased to 200, the semantic segmentation performance of the GraSS still obtains an improvement, although it fluctuates slightly. Among them, the semantic segmentation performance of the LoveDA Urban dataset fluctuates significantly, and for the Potsdam and LoveDA Rural datasets, the best semantic segmentation OA was obtained when the warm-up epoch is 200.

\paragraph{Analysis of the Threshold of the Activation Map for Selecting DAR}

To analyze the effect of the threshold $T_A$ of the activation map for selecting DAR on the semantic segmentation performance of the model, we selected five thresholds of 0, 0.3, 0.5, 0.7 and 0.9 for comparison experiments on the Potsdam, LoveDA Urban, and LoveDA Rural datasets.

The threshold of 0 indicates that no gradient guided sampling is performed, and a larger threshold indicates that a smaller sampling region is obtained. The results in Fig. \ref{fig_ta} show that gradient guided sampling effectively improves the semantic segmentation performance of the contrastive learning model and achieves the best semantic segmentation performance at a threshold value of 0.5. However, too large threshold $T_A$ will cause the model to select too small RSI regions, resulting in the constructed samples containing too little or missing ground object information, which causes a degradation of the semantic segmentation performance.

\subsubsection{Analysis of the Number of Ground Objects Contained in the Sample}

To analyze the number of ground objects contained in the samples obtained by GraSS, we use the label information to count the number of ground objects contained in the sample by gradient guided sampling strategy (GraSS) on Potsdam, LoveDA Urban, and LoveDA Rural datasets and compared it with the original RSI sample and random resize crop. In this analysis experiment, the batch size was set to 32, and the gradient guided sampling training was started after 150 epochs of instance discrimination warm-up and continued to be observed for 50 epochs. The experimental results are shown in Fig. \ref{fig_num}.

We first analyzed the number of ground objects contained in the average single sample in the first batch of each training epoch, and the experimental results are shown in the first column of Fig. \ref{fig_num}. The experimental results show that the samples obtained by GraSS contain the lowest number of ground objects compared to the original RSI and random resize crop in the 50 observed epochs, which indicates that GraSS effectively reduces the number of ground objects contained in the original RSI samples.

In addition, we also analyzed the number of samples containing single ground objects in the first batch of each training epoch, and the experimental results are shown in the second column of Fig. \ref{fig_num}. The experimental results show that GraSS obtains the highest number of samples containing single ground objects compared to the original RSI and the random resize crop in the 50 observed epochs. For the Potsdam dataset, GraSS obtained a maximum of 19 samples containing single ground objects compared to the original RSI, for the LoveDA Urban dataset, GraSS obtained a maximum of 18 samples containing single ground objects compared to the original RSI, and for the LoveDA Rural dataset, GraSS obtained a maximum of 23 samples containing single ground objects compared to the original RSI. This indicates that GraSS can obtain more samples containing single ground objects, effectively mitigating positive sample confounding issue and feature adaptation bias.

\subsubsection{Visual analysis of contrastive Loss Attention Map (LAM)}

\begin{figure*}[ht]
\vspace{-5mm}
\centering
\includegraphics[width=1\linewidth]{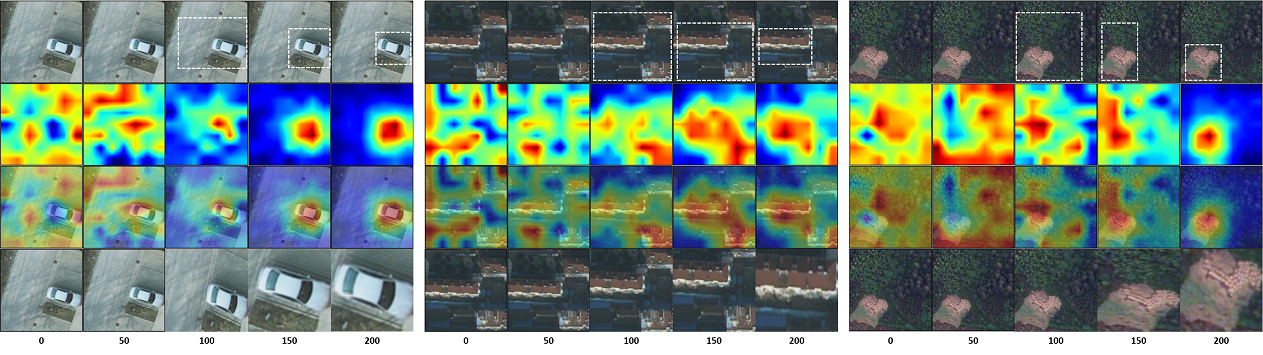}
\caption{The result of the effect of instance discrimination warm-up Epoch on LAM. The first row indicates the original RSI, the white dashed box is the cropped area obtained when the threshold $T_A$ is set to 0.5, the second row is the LAM corresponding to the RSI, the third row shows the superimposed results of the RSI and the corresponding LAM, the fourth row is the samples reconstructed from GraSS after instance discrimination warm-up, and the number in the last row indicates the corresponding instance discrimination warm-up epoch.}
\label{fig_ea}
\end{figure*}

\begin{figure}[htp]
\centering
\includegraphics[width=1\linewidth]{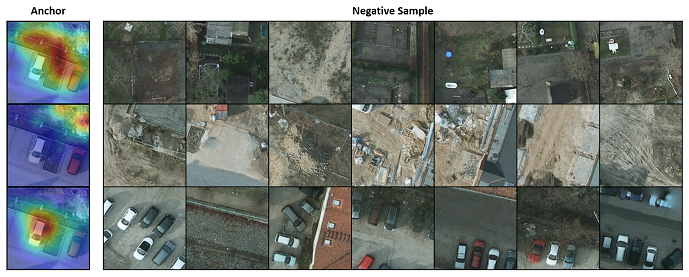}
\includegraphics[width=1\linewidth]{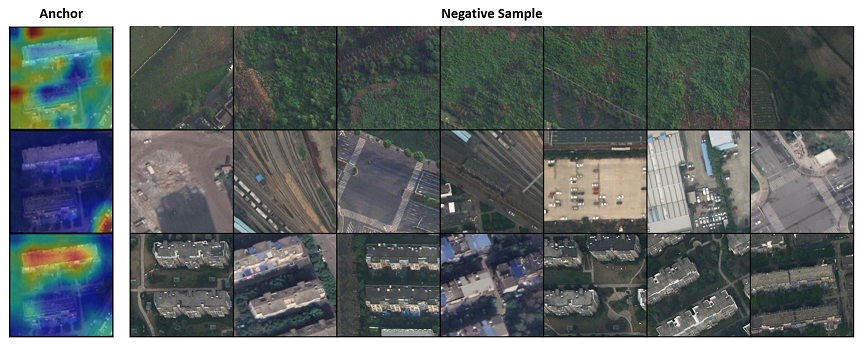}
\caption{The experiment result of the effect of ground objects contained in RSI on LAM. Each row represents a batch of remote sensing images that are input to the model. Among them, the anchor samples in the first three rows are the same, and the negative samples are different. The anchor samples in the last three rows are the same, and the negative samples are different.}
\label{fig_n}
\end{figure}

\paragraph{Effect of Instance Discrimination Warm-Up Epoch on the LAM}

To explore the effect of instance discrimination warm-up epoch on the contrastive Loss Activation Map (LAM), we visualized the LAM obtained from different instance discrimination warm-up epochs, and the experimental results are shown in Fig. \ref{fig_ea}.

We selected 5 instance discrimination warm-up epochs of 0, 50, 100, 150, and 200 for visualization and analysis of the LAM. The experimental results show that as the instance discrimination warm-up proceeds, the obtained LAMs gradually focus on a certain region of the RSI, which tends to contain more singular ground objects in RSI.

\paragraph{Effect of Ground Objects Contained in RSI on the LAM}

To explore the effect of ground objects contained in RSI on LAM, we artificially constructed different batches of image data and acquired LAM for visualization, and the batch size is set to 8. Specifically, we first specify the anchor sample images and observe the changes in the anchor sample's LAM by replacing other images in the same batch. Where other images in the same batch are considered as negative samples by self-supervised contrastive learning, the experimental results are shown in Fig. \ref{fig_n}.

The experimental results show that the regions with higher activation values in LAM tend to be concentrated on a relatively large number of ground objects in the same batch. As shown in the first to third rows of Fig. \ref{fig_n}, for the same anchor sample image, when other images in the same batch contain a high number of grass or low vegetation, the regions with higher activation values in the anchor sample's LAM are concentrated in the low vegetation area (as shown in the first row of Fig. \ref{fig_n}). When other images in the same batch contain a high number of clutter, the regions with higher activation values in the anchor sample's LAM are concentrated in the clutter area (as shown in the second row of Fig. \ref{fig_n}). When other images in the same batch contain a high number of cars, the regions with higher activation values in the anchor sample's LAM are concentrated in the car area (as shown in the third row of Fig. \ref{fig_n}). And the fourth to sixth rows of Figure 9 show similar results.

A possible explanation for such a result is that when ground objects in the negative sample are close to those in the anchor sample, a larger contrastive loss gradient will be generated, which leads to a higher activation value of the image area corresponding to the LAM obtained from the contrastive loss gradients.

\section{Conclusion}

In this paper, we propose contrastive learning with Gradient guided Sampling Strategy (GraSS) for RSI semantic segmentation. It uses the positive and negative sample discrimination information contained in the self-supervised contrastive loss gradients to construct samples containing more singular ground objects, alleviate the sample confounding issue of semantic segmentation of RSIs for self-supervised contrastive learning, and mitigate the feature adaptation bias between instance-level pretext task and pixel-level RSI semantic segmentation tasks. The experiments show that the GraSS effectively improves the performance of the self-supervised contrastive learning model for the RSI semantic segmentation task and outperforms a total of eight self-supervised contrastive learning methods of six types at present. In addition, we have conducted extensive experiments and probes on the proposed GraSS and preliminarily discussed the effects of two factors, the instance discrimination warm-up epoch and the ground objects contained in RSI, on the LAM obtained by contrastive loss gradients, which is expected to deepen our understanding of self-supervised contrastive learning models.

Although the current gradient guided sampling strategy effectively mitigates the positive sample confounding issue of self-supervised contrastive learning for the RSI semantic segmentation task, since self-supervised contrastive learning is essentially unsupervised, our proposed GraSS cannot absolutely guarantee that the obtained sample contains only a single type of ground objects.

In addition, we found that the contrastive loss gradient contains rich feature information, which inspires us to make more use of the gradient information in the process of model training to obtain additional model capabilities. However, we currently lack a clear understanding of several factors that affect the contrastive Loss Attention Map (LAM) obtained by the contrastive loss gradients. In the future, we will further explore the relationship between the contrastive loss gradient and the spatio-temporal characteristics of RSI, which may provide guidance for designing a self-supervised contrastive learning model that can capture the features of RSIs more effectively.

\vspace{-5pt}
\section*{Acknowledgment}

The authors would like to thank the valuable comments from anonymous reviewers.

\bigskip
\bibliographystyle{IEEEtran} 
\bibliography{GraSS_paper} 

\vspace{-20pt}

\end{document}